\documentclass[conference,letterpaper]{IEEEtran}
\IEEEoverridecommandlockouts
\usepackage[T1]{fontenc}
\usepackage{cite}
\usepackage{graphicx,amsmath,amssymb,booktabs,array,multirow,xcolor,xspace}
\usepackage{microtype,placeins}
\usepackage{capt-of}
\usepackage[hidelinks,hypertexnames=false]{hyperref}
\usepackage[capitalise,noabbrev]{cleveref}

\newcommand{\src}[1]{}

\definecolor{scoregreen}{RGB}{25,125,72}
\newcommand{\bestscore}[1]{\textcolor{scoregreen}{\textbf{#1}}}

\hypersetup{pdftitle={MoRE: Motion-Regularized Encoding for Articulated 3D Generation},pdfauthor={Guantian Zheng, Shan Jin, Tim Tso, Hongfei Cai}}
\renewcommand{\footnoterule}{\kern-3pt\hrule width 0.4\columnwidth height 0.4pt\kern2.6pt}
\title{MoRE: Motion-Regularized Encoding for Articulated 3D Generation}
\author{Guantian Zheng\textsuperscript{1,3,*},
Shan Jin\textsuperscript{1,2,*},
Tim Tso\textsuperscript{1,2,*},
Hongfei Cai\textsuperscript{1,2,*}\\
{\small \textsuperscript{1}Manifest Limited\quad
\textsuperscript{2}Department of Engineering, University of Cambridge}\\
{\small \textsuperscript{3}School of Mechanical and Aerospace Engineering, Nanyang Technological University}
}

\begin{document}
\maketitle
\suppressfloats[t]
\begingroup
\renewcommand{\thefootnote}{\fnsymbol{footnote}}
\footnotetext[1]{These authors contributed equally.}
\footnotetext[0]{Corresponding author: \href{mailto:gzheng004@e.ntu.edu.sg}{\nolinkurl{gzheng004@e.ntu.edu.sg}}.}
\endgroup
\begin{abstract}
Generating an articulated object from a single image requires recovering both its appearance and the mechanism governing its parts.
A compact encoding of joint attributes does not by itself ensure that reconstruction preserves the resulting motion.
We present MoRE, Motion-Regularized Encoding for articulated 3D generation.
MoRE represents coordinate-aligned joint lines with axis classification and two axis-orthogonal Pl\"ucker moment coefficients, removing the arbitrary choice of an origin along the axis.
During representation learning, decoded joint hypotheses are used to move corresponding three-dimensional anchors, and their motions are compared with those of annotated joints.
A differentiable expectation over joint types and axis directions couples attribute reconstruction to the motion it induces.
An image-conditioned flow learns the resulting functional latents, while a multi-state-adapted Gaussian model supplies appearance.
At inference, recovered part assignments and joints articulate a single set of rest-pose Gaussians without regenerating the object for each state.
Experiments on PartNet-Mobility and ACD show stronger part-box agreement, articulated-state geometry and visual similarity than the compared baselines, while joint prediction reduces axis, origin and range errors relative to ArtiLatent.
Controlled ablations support the joint-line encoding and show that motion supervision reduces anchor-motion error.
Code and pretrained checkpoints will be released in the future.
\end{abstract}

\section{Introduction}
Image-conditioned 3D generation has advanced in geometry and appearance, but an articulated object must also move correctly.
A door can look plausible while rotating about the wrong hinge; a drawer can reconstruct well in its closed state while translating along an incorrect direction.
Recovering static geometry and reconstructing joint attributes are therefore insufficient tests of an articulated representation.

\begin{figure}[t]
\centering
\includegraphics[width=\linewidth]{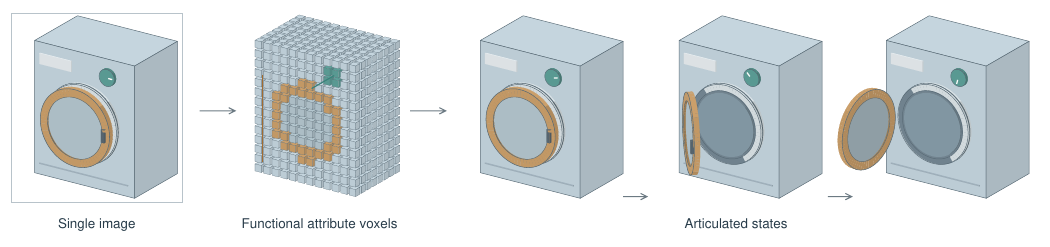}
\caption{From a single image to functional voxels and articulated configurations. Earlier attribute-only checkpoint.}
\label{fig:icme_teaser}
\end{figure}

Recent methods generate articulated objects through part graphs, meshes and structured latents~\cite{liu2025singapo,gao2025meshart,chen2025artilatent,liu2026pact}.
A central challenge is to connect the representation of a joint to the motion it induces.
A point on an axis is not unique: moving it along the axis leaves rotation unchanged.
A joint-line representation removes this redundancy, but attribute-wise reconstruction still treats errors separately rather than measuring their combined effect on moving geometry.

We introduce MoRE, \emph{Motion-Regularized Encoding}, which learns sparse functional latents under both attribute and analytic motion supervision.
For coordinate-aligned joints, axis classification and two orthogonal Pl\"ucker moment coefficients specify the joint line.
Decoded attributes are pooled within training parts to form soft joint hypotheses.
At shared articulation progress values, predicted and annotated joints move the same fixed anchors; the expected squared discrepancy regularizes the encoding.
This objective measures motion before choosing a hard joint type or axis, allowing gradients to reach both classification and continuous predictions.

An image-conditioned flow learns the motion-regularized latents.
At inference, predicted part assignments connect recovered joints to a multi-state-adapted Gaussian appearance model.
The object is decoded once and articulated through rigid part transforms, without ground-truth joints or part masks.
Our contributions are:
\begin{itemize}
\item A differentiable, part-balanced motion regularizer that supervises the behavior of decoded joint hypotheses in addition to their individual attributes.
\item An axis-conditioned functional encoding that specifies coordinate-aligned joint lines without an arbitrary along-axis origin.
\item A unified single-image generation pipeline with shared visual conditioning and part-bound Gaussian appearance, allowing the same decoded object to be articulated across states.
\end{itemize}

\section{Related Work}
\paragraph{Image-conditioned 3D generation}
Single-image 3D generation learns priors to infer geometry and appearance beyond the visible surfaces.
Structured latent models associate local features with spatial support, allowing detailed geometry and appearance to be decoded from a compact representation.
TRELLIS~\cite{xiang2025trellis} uses sparse voxel latents with multiple output decoders, while TRELLIS.2~\cite{xiang2026trellis2} models geometry and material attributes with O-Voxels.
These static representations provide useful spatial and appearance priors, but do not specify the parts and joints needed for articulation.
MoRE associates functional attributes with sparse support and connects the resulting parts to Gaussian appearance.

\paragraph{Articulated object generation}
NAP~\cite{lei2023nap} models articulated objects through part geometry and joint relations, and CAGE~\cite{liu2024cage} adds control over articulation structure.
SINGAPO~\cite{liu2025singapo} conditions part attributes on an image and a connectivity graph before retrieving meshes.
MeshArt~\cite{gao2025meshart} generates meshes hierarchically, while ArtFormer~\cite{su2025artformer} generates tokenized kinematic trees.
More recent methods combine pretrained 3D priors with articulation: PAct~\cite{liu2026pact} generates part-decomposed latents and predicts joints from part features; ArtiLatent~\cite{chen2025artilatent} jointly encodes voxel geometry and joint attributes and adapts Gaussian appearance across states.
PhysX-Anything~\cite{cao2025physxanything} additionally models physical properties, while MonoArt~\cite{li2026monoart} progressively reasons over geometry, parts and motion.
MoRE studies motion-regularized encoding: joint-line attributes are supervised together with the analytic motion they induce.
Pl\"ucker joint representations also appear in NAP~\cite{lei2023nap}; here, a classified coordinate axis determines an orthogonal plane in which two moment coefficients specify the line.

\paragraph{Articulated reconstruction and appearance}
Reconstruction methods recover geometry and motion from observations of the same object in different states.
PARIS~\cite{liu2023paris} jointly estimates part-level implicit geometry, appearance and motion from multi-view images in two articulation states.
ArticulatedGS~\cite{guo2025articulatedgs} uses a Gaussian representation and progressive optimization to recover parts and motion from such observations.
These approaches exploit observed state changes, whereas single-image generation must infer unobserved structure from learned priors.
MoRE uses multi-state observations to adapt its appearance model during training, and associates rest-pose Gaussians with explicit joints through part assignments.

\section{Method}
\begin{figure*}[t]
\centering
\includegraphics[width=\textwidth]{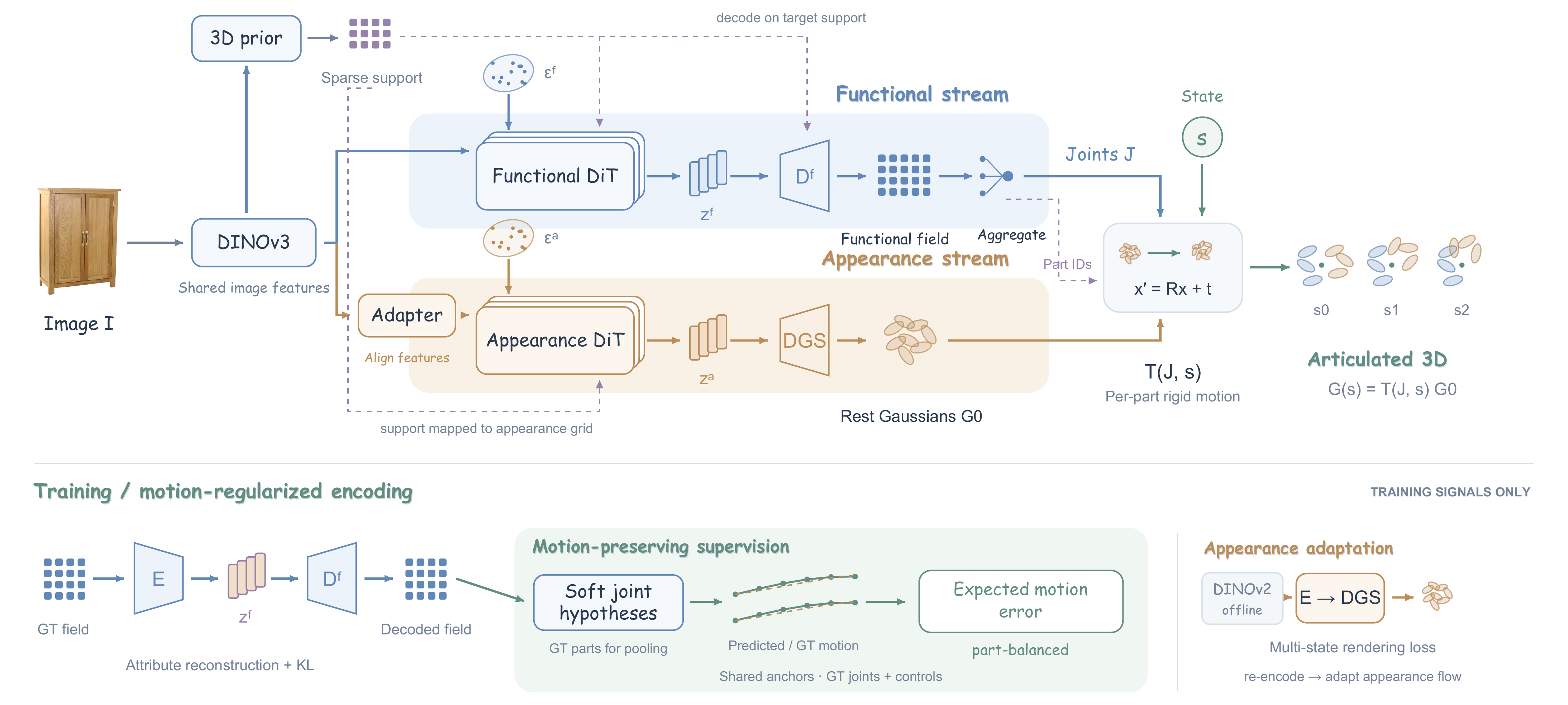}
\caption{\textbf{MoRE overview.} A shared DINOv3 conditions sparse support, functional and appearance generation; an adapter conditions the appearance branch. Training adds expected anchor-motion error to attribute reconstruction; appearance is adapted across states. Cabinet photo: illustrative input, not a tested generation result; PNGimg~\cite{pngimgCabinet}, CC BY-NC 4.0.}
\label{fig:icme_pipeline}
\end{figure*}

Given an object image $I$, MoRE generates an articulated representation $\mathcal O=(\mathcal V,\mathcal P,\mathcal J,\mathcal G_0)$ comprising sparse support, part assignments, part-level joints and rest-pose Gaussians.
The functional stream predicts part and motion attributes; the appearance stream provides renderable geometry and color.
Part assignments and joint transforms connect the streams (\cref{fig:icme_pipeline}).

\subsection{Functional voxel representation}

\paragraph{Voxel attributes}
On rest-pose sparse support $\mathcal V=\{x_i\}_{i=1}^{N}$, each voxel stores semantic/bounding-box descriptors, joint type, axis class, moment and limits.
Descriptors separate part instances; joint attributes describe their shared motion.
Density and friction are metadata, not inputs to articulation.
\paragraph{Axis-orthogonal joint lines}
An axis point is not unique: translating it along the axis does not change the joint.
For a unit direction $\ell$ through $o$, the Pl\"ucker moment $m=o\times\ell$ is independent of this translation and satisfies $m\perp\ell$.
For coordinate-aligned joints, we classify $\ell\in\{e_x,e_y,e_z\}$ and predict two coefficients:
\begin{equation}
m=c_1u_\ell+c_2v_\ell,\qquad o_\perp=\ell\times m,
\end{equation}
where $(u_\ell,v_\ell)$ span the selected axis's orthogonal plane and $o_\perp$ is the closest point on the line to the origin.
This specifies the joint line without regressing a redundant along-axis coordinate.
Reversing a signed axis requires reversing both its moment and joint displacement to preserve the line and motion. The three axes use the orthogonal bases $(e_y,e_z)$, $(e_z,e_x)$ and $(e_x,e_y)$, respectively.

\subsection{Functional latent generation}
A sparse articulation VAE~\cite{kingma2014vae} compresses the $96^3$ field to $24^3$ with 16 latent channels.
The image-conditioned flow is initialized from a compatible structure-flow checkpoint.

\paragraph{Learning functional latents}
The encoder predicts a diagonal Gaussian posterior, and the decoder reconstructs voxel attributes from a sampled latent $z$.
The VAE objective is
\begin{equation}
\begin{aligned}
\mathcal L_{\mathrm{VAE}}={}&\mathcal L_{\mathrm{attr}}
+\beta \overline D_{\mathrm{KL}}(q_\phi\|\mathcal N(0,I))\\
&+\lambda_M\mathcal L_{\mathrm{motion}}.
\end{aligned}
\end{equation}
The reconstruction term combines classification losses for semantic labels, joint types and axis classes with regression losses for bounding boxes, moments and motion limits.
Axis supervision is restricted to movable voxels, since fixed parts have no motion axis.
Here $q_\phi$ is the encoder posterior, $\beta$ the KL weight, and $\mathcal L_{\mathrm{attr}}$ the weighted attribute reconstruction loss. $\overline D_{\mathrm{KL}}$ is the KL divergence averaged over latent tokens and channels, matching the training reduction.

\paragraph{Motion-regularized encoding}
During VAE training, we average raw decoder predictions within each ground-truth part.
Softmax probabilities define joint-type and axis hypotheses $c$, with weights $\pi_{p,c}$; no hard voting is used in this loss. The fixed-hypothesis weight is $\pi_{p,0}=p_p(\mathrm{fixed})$; for moving type $\tau$ and axis $a$, $\pi_{p,(\tau,a)}=p_p(\tau)p_p(a)$. Type probabilities are normalized over fixed, revolute and prismatic classes. The spherical class in the stored layout is outside this single-DoF motion term and retains type-classification supervision.
Each movable training part $p$ supplies fixed voxel-center anchors $\mathcal A_p$, deduplicated and deterministically subsampled to at most 256 points.
For normalized progress $u\in\mathcal U=\{0,0.5,1\}$, a joint $J$ uses its own physical limits and the annotated reference coordinate:
\begin{equation}
q(J,u)=r^-+u(r^+-r^-)-q_{\mathrm{ref}}.
\label{eq:motion_progress}
\end{equation}
Predicted and annotated limits are interpolated separately at the same $u$; the target motion therefore remains fixed if the predicted range shrinks.
Revolute displacement is measured in radians and prismatic displacement in normalized object units.
With $J_p^*$ the annotated joint and $T$ the analytic rigid transform, we minimize
\begin{equation}
\begin{aligned}
\ell_p={}&\sum_c\pi_{p,c}\frac{1}{|\mathcal U||\mathcal A_p|}
\sum_{u\in\mathcal U}\sum_{x\in\mathcal A_p}\\[-2pt]
&\left\|T(\hat J_{p,c},u)x-T(J_p^*,u)x\right\|_2^2,\\
\mathcal L_{\mathrm{motion}}={}&\frac{1}{|\mathcal B^+|}
\sum_{b\in\mathcal B^+}\frac{1}{|\mathcal P_b^+|}
\sum_{p\in\mathcal P_b^+}\ell_p.
\end{aligned}
\label{eq:motion_loss}
\end{equation}
Here $\mathcal P_b^+$ contains movable parts and $\mathcal B^+$ contains objects with such parts.
Errors are squared before averaging over hypotheses, preventing opposite motions from canceling. If $\mathcal B^+=\varnothing$, the motion loss is zero.
The fixed hypothesis contributes identity motion; existing type classification supervises fixed parts.
The expectation propagates gradients to type/axis logits, moment coefficients and motion limits.
Part and object means balance voxel counts, without per-part size normalization.
GT parts supervise training only; inference clusters predictions.
The objective evaluates independent part motion, not dynamics or hierarchical kinematics.

\paragraph{Image-conditioned flow}
After VAE training, we freeze it and normalize the encoded latents channel-wise.
Let $z_0$ be a clean latent, $\epsilon\sim\mathcal N(0,I)$, and $t\in[0,1]$.
Following flow matching and rectified flow~\cite{lipman2023flowmatching,liu2023rectifiedflow}, with a small noise floor $\delta$, the interpolation and velocity target are
\begin{equation}
\begin{aligned}
z_t&=(1-t)z_0+[\delta+(1-\delta)t]\epsilon,\\
v^*&=(1-\delta)\epsilon-z_0.
\end{aligned}
\end{equation}
A transformer $v_\theta$~\cite{peebles2023dit} receives sparse latent tokens, time and image features $h_I$ and minimizes
\begin{equation}
\mathcal L_{\mathrm{flow}}=\mathbb E\|v_\theta(z_t,t,h_I)-v^*\|_2^2.
\end{equation}
Image features enter through cross-attention, while self-attention relates spatial tokens.
Sampling integrates the learned velocity from noise at $t=1$ toward the near-clean endpoint at $t=0$, where the training path is $z_0+\delta\epsilon$. The decoder then recovers the functional field on the generated support.

\paragraph{Recovering parts and joints}
We first group voxels by predicted semantic class and apply DBSCAN~\cite{ester1996dbscan} to their bounding-box descriptors within each group.
The resulting labels define voxel sets $\mathcal P_k$ for part instances, determining the part count through clustering.
For each part, majority votes select joint type and axis; moments and limits are averaged within the part.
The aggregated axis selects the orthogonal basis used to recover its joint line.
This yields a joint $J_k=(\tau_k,\ell_k,o_k,r_k)$ and a single rigid motion shared by the part's voxels.
The embedding-based configuration used in the functional evaluation is specified separately in the experimental setup.
\begin{table*}[t]
\centering
\caption{Single-image articulated generation. Green marks the best values on each dataset, including ties.}
\label{tab:icme_benchmark}
\footnotesize\renewcommand{\arraystretch}{1.08}
\setlength{\tabcolsep}{3pt}
\begin{tabular*}{\textwidth}{@{\extracolsep{\fill}}lcccccc@{}}
\toprule
Method & RS-$d_{\mathrm{gIoU}}\downarrow$ & AS-$d_{\mathrm{gIoU}}\downarrow$ & RS-CD$\downarrow$ & AS-CD$\downarrow$ & AOR$\downarrow$ & CLIP$\uparrow$ \\
\midrule
\multicolumn{7}{@{}l}{\textbf{PartNet-Mobility}} \\
SINGAPO~\cite{liu2025singapo} & 0.5119 & 0.5166 & 0.0040 & 0.0202 & 0.0097 & 0.8779 \\
ArtFormer~\cite{su2025artformer} & 1.3165 & 1.3212 & 0.0504 & 0.1849 & 0.0191 & 0.7877 \\
PhysX-3D~\cite{cao2025physx} & 0.7610 & 0.7637 & 0.0115 & 0.1204 & 0.2125 & 0.7972 \\
PhysX-Anything~\cite{cao2025physxanything} & 0.9596 & 0.9661 & 0.0097 & 0.1106 & \bestscore{0.0056} & 0.8859 \\
Articulate-Anything~\cite{le2025articulateanything} & 0.6865 & 0.6969 & 0.0107 & 0.1230 & \bestscore{0.0056} & 0.8795 \\
PAct (SAM+VLM)~\cite{liu2026pact} & 0.2485 & 0.2736 & \bestscore{0.0019} & 0.0176 & 0.0233 & 0.8996 \\
\textbf{MoRE (Ours)} & \bestscore{0.2041} & \bestscore{0.11834} & 0.0028 & \bestscore{0.0092} & 0.0101 & \bestscore{0.9545} \\
\midrule
\multicolumn{7}{@{}l}{\textbf{ACD}} \\
SINGAPO~\cite{liu2025singapo} & 0.7272 & 0.7306 & 0.0090 & 0.0256 & 0.0115 & 0.8376 \\
ArtFormer~\cite{su2025artformer} & 1.3489 & 1.3876 & 0.0396 & 0.0825 & 0.0147 & 0.7690 \\
PhysX-3D~\cite{cao2025physx} & 0.6449 & 0.6464 & 0.0209 & 0.0677 & 0.0960 & 0.8132 \\
PhysX-Anything~\cite{cao2025physxanything} & 1.1673 & 1.1709 & 0.0162 & 0.0481 & 0.0099 & 0.8456 \\
Articulate-Anything~\cite{le2025articulateanything} & 1.2677 & 1.2693 & 0.2583 & 0.2803 & 0.1236 & 0.8648 \\
PAct (SAM+VLM)~\cite{liu2026pact} & 0.5316 & 0.5412 & \bestscore{0.0065} & 0.0218 & 0.0248 & 0.8901 \\
\textbf{MoRE (Ours)} & \bestscore{0.4714} & \bestscore{0.4783} & 0.0097 & \bestscore{0.0145} & \bestscore{0.0087} & \bestscore{0.9401} \\
\bottomrule
\end{tabular*}
\par\smallskip
{\footnotesize\raggedright Baseline scores are from PAct~\cite{liu2026pact} Tables 1--2. PAct uses automatic SAM+VLM masks; SINGAPO uses GT connectivity graphs; ArtFormer uses image-to-text conditioning. RS/AS: rest/articulated states.\par}
\end{table*}

\begin{table*}[t]
\centering
\caption{Joint prediction on PartNet-Mobility. Green marks the best values; baseline scores are from ArtiLatent Table 3.}
\label{tab:icme_functional}
\small\renewcommand{\arraystretch}{1.1}
\setlength{\tabcolsep}{4pt}
\begin{tabular*}{\textwidth}{@{\extracolsep{\fill}}lccccc@{}}
\toprule
Method & Type (\%)$\uparrow$ & Axis err. ($^\circ$)$\downarrow$ & Origin$\downarrow$ & Range (rot.) ($^\circ$)$\downarrow$ & Range (trans.)$\downarrow$ \\
\midrule
SINGAPO~\cite{liu2025singapo} & 97.37 & 1.29 & 0.39 & 6.73 & 0.0265 \\
ArtiLatent~\cite{chen2025artilatent} & \bestscore{99.17} & 1.14 & 0.10 & 7.37 & 0.0159 \\
\textbf{MoRE} & 99.16 & \bestscore{1.03} & \bestscore{0.04} & \bestscore{6.29} & \bestscore{0.0107} \\
\bottomrule
\end{tabular*}
\par\smallskip
{\footnotesize\raggedright Metrics retain the source definitions; origin error is not axis-line distance.\par}
\end{table*}

\begin{table*}[t]
\centering
\caption{Joint-line encoding and motion regularization on PartNet-Mobility. Green marks the best values.}
\label{tab:icme_motion_ablation}
\footnotesize\renewcommand{\arraystretch}{1.08}
\setlength{\tabcolsep}{3pt}
\begin{tabular*}{\textwidth}{@{\extracolsep{\fill}}llcccccc@{}}
\toprule
Encoding & $\mathcal L_{\mathrm{motion}}$ & Axis err. ($^\circ$)$\downarrow$ & Line dist.$\downarrow$ & RMSE-R$\downarrow$ & RMSE-P$\downarrow$ & Range-R ($^\circ$)$\downarrow$ & Range-P$\downarrow$ \\
\midrule
Axis + origin & off & 1.24 & 0.0253 & 0.0477 & 0.0214 & 7.28 & 0.0132 \\
Axis + moment & off & 1.17 & 0.0229 & 0.0431 & 0.0207 & 7.20 & 0.0123 \\
Axis + origin & on & 1.15 & 0.0206 & 0.0383 & 0.0185 & 6.31 & 0.0117 \\
\textbf{Axis + moment (MoRE)} & on & \bestscore{1.03} & \bestscore{0.0191} & \bestscore{0.0362} & \bestscore{0.0168} & \bestscore{6.29} & \bestscore{0.0107} \\
\bottomrule
\end{tabular*}
\par\smallskip
{\footnotesize\raggedright Full-MoRE axis and range errors correspond to Table~\ref{tab:icme_functional}; variants share the axis classifier, inputs, test set, part aggregation and statistics. off/on disables/enables the motion loss; R/P: revolute/prismatic. Axis and rotational range errors are in degrees, other distances in normalized object units. Line dist. is the shortest distance between revolute-axis lines, distinct from Origin; RMSE-R/P measure anchor-motion errors for ground-truth revolute/prismatic parts, respectively.\par}
\end{table*}

\subsection{Articulation-aware appearance}
Opening a part exposes interior surfaces absent from a closed-state image.
We therefore adapt the pretrained TRELLIS~\cite{xiang2025trellis} appearance encoder $E_A$ and Gaussian decoder $D_A$ using multi-view observations across articulation states, a strategy also studied in articulated generation~\cite{chen2025artilatent}.
Offline DINOv2 features~\cite{oquab2024dinov2} are projected and aggregated onto rest-pose voxels, then encoded into appearance latents $z_A$.
The decoder predicts 32 Gaussians~\cite{kerbl2023gaussians} per active appearance voxel.
Each Gaussian inherits its voxel's part assignment, providing an explicit correspondence between appearance and joint motion.

For a training state $s$, known joints transform decoded Gaussians before rendering $\hat I_s$.
We optimize
\begin{equation}
\begin{aligned}
\mathcal L_A=\mathbb E_s[&0.8\mathcal L_1(\hat I_s,I_s)\\
&+0.2\mathcal L_{\mathrm{D\text{-}SSIM}}(\hat I_s,I_s)]
+\lambda_R\mathcal L_R.
\end{aligned}
\end{equation}
Here $\mathcal L_{\mathrm{D\text{-}SSIM}}=1-\mathrm{SSIM}$~\cite{wang2004ssim} and $\mathcal L_R$ penalizes the mean Gaussian scale.
Supervising multiple states exposes the decoder to both exterior and motion-revealed surfaces.
After adapting $E_A,D_A$, we re-encode training objects and fine-tune the appearance flow to the updated latent distribution.
At inference, a shared DINOv3~\cite{simeoni2026dinov3} conditions all branches; a learnable adapter supplies $h^a=A_\psi(E_v(I))$ as appearance-flow conditions. The encoder $E_v$ is frozen; the adapter is optimized with the appearance flow on re-encoded targets.
DINOv2 is used only offline to construct appearance training latents.
\subsection{Part-wise motion and inference}
Let $q_k$ denote joint motion relative to the rest pose: radians for revolute joints and distance in normalized object coordinates for prismatic joints. For Gaussian $g$ assigned to part $k$, its center follows
\begin{equation}
x_g'=\begin{cases}
R(\ell_k,q_k)(x_g-o_k)+o_k,&\text{revolute},\\
x_g+q_k\ell_k,&\text{prismatic}.
\end{cases}
\end{equation}
Fixed parts remain unchanged. Generation uses prescribed displacements relative to the decoded rest pose, without an annotated reference coordinate.
These transforms assume joint parameters and Gaussian centers share the same object coordinate frame.
Gaussian orientation follows the same rotation; color, scale and opacity are reused across states.
Part assignments are transferred from the $96^3$ functional grid to the $64^3$ appearance grid.

\paragraph{Inference}
Given an image, we generate sparse support, then sample and decode functional and appearance latents.
Part clustering recovers explicit joints and assignments, which are transferred to appearance voxels and their Gaussians.
For a prescribed joint state, rigidly transforming each part's Gaussians renders a new configuration without regenerating the functional field or appearance.
VAE encoders and multi-state observations are used during training; image-conditioned flows supply the latents during generation.

\section{Experiments}
\subsection{Comparison with prior methods}
\Cref{tab:icme_benchmark} compares MoRE with prior methods on PartNet-Mobility~\cite{xiang2020sapien} and ACD~\cite{iliash2026s2o}, using baseline scores reported by PAct~\cite{liu2026pact}.
\paragraph{Single-image generation} MoRE ranks first on four of six metrics on PartNet-Mobility and five on ACD.
Compared with PAct, AS-CD decreases from 0.0176 to 0.0092 and from 0.0218 to 0.0145, respectively; CLIP reaches 0.9545 and 0.9401.
However, RS-CD remains higher than PAct on both datasets, and PartNet-Mobility AOR is not the best.
These results show stronger articulated-state geometry, while rest-state surface reconstruction remains a limitation.
\paragraph{Joint prediction} \Cref{tab:icme_functional} compares MoRE with the joint results reported by ArtiLatent~\cite{chen2025artilatent}.
MoRE has lower axis, origin and both range errors: axis error decreases from $1.14^\circ$ to $1.03^\circ$, and origin error from 0.10 to 0.04.
Type accuracy is 99.16\%, 0.01 percentage points below ArtiLatent's 99.17\%.
These results support joint prediction by the complete model; the controlled ablations below further examine encoding and motion supervision.

\begin{figure}[t]
\centering
\includegraphics[width=\linewidth]{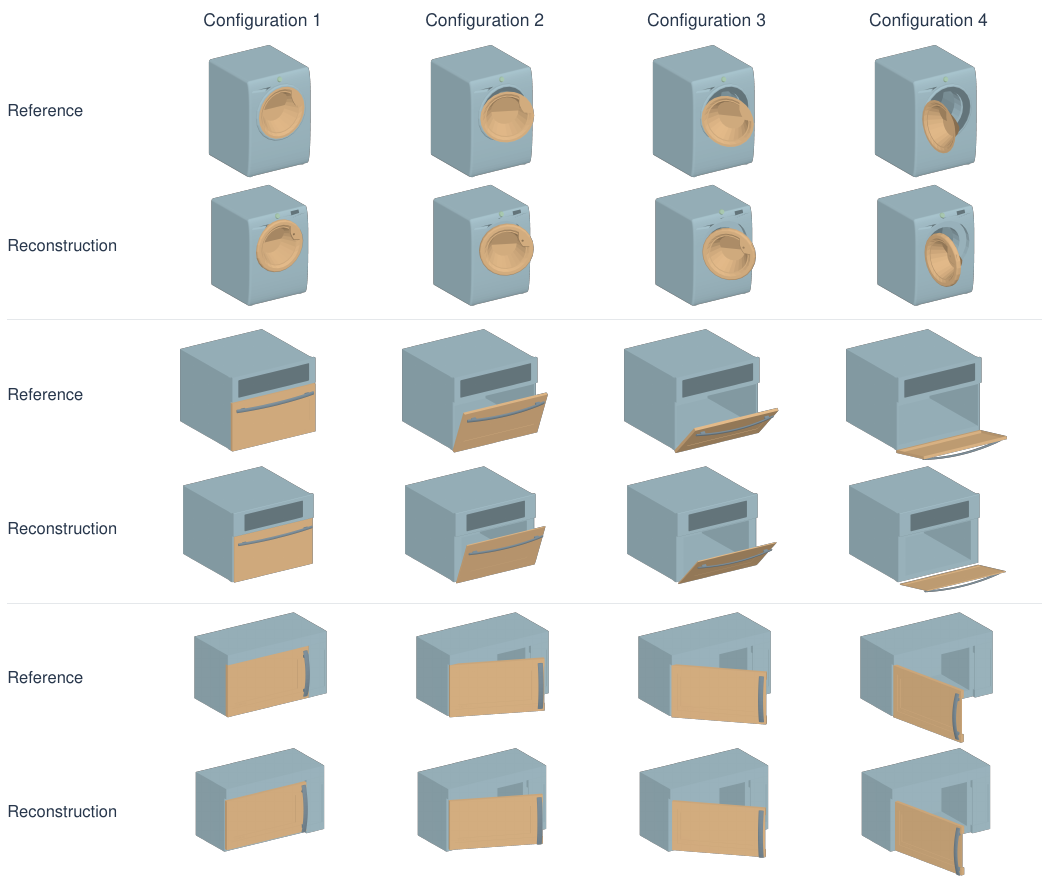}
\caption{Three objects across four configurations, with reference above reconstruction in each pair. Known joints articulate Gaussians decoded from encoder latents.}
\label{fig:icme_states}
\end{figure}

\subsection{Experimental setup}
Public benchmarks evaluate single-image generation using rest/articulated-state (RS/AS) part-box distance, surface CD, part overlap and CLIP similarity. Axis in the joint and ablation tables is angular error in degrees; the earlier component axis-classification accuracy below is a different metric in percent. The shape-conditioned and appearance-reconstruction settings below concern earlier component experiments, separate from the main tables.
Functional learning uses 499,329 procedural objects across 18 categories from Infinigen-Articulated~\cite{joshi2025infinigenarticulated}.
The earlier shape-conditioned evaluation uses 144 held-out objects stratified by topology, with no functional ground truth supplied.
The earlier functional evaluation uses a 32-channel shape latent and a 32-channel noisy functional latent with DINOv3 conditioning.
A 16-dimensional contrastive part embedding~\cite{khosla2020supcon} replaces semantic/bounding-box descriptors; spherical $k$-means~\cite{dhillon2001sphericalkmeans} uses a supplied part count $K$.
Type and axis use modes, and continuous parameters use coordinate-wise medians.
Sampling uses 25 Euler steps and guidance scale 1.0.

Appearance adaptation uses a 3,276-object SINGAPO-derived collection~\cite{liu2025singapo}, rendered in eight states from 48 views.
Reconstruction measurements use asset joints and encoder latents, not generated articulation and appearance latents.
The renderer samples $q_k(s)=(r_k^+-r_k^-)s/S$ with $S=8$, reaching $7/8$ of the supplied travel in the last state.
The teaser and qualitative image-only generation use image input; they are separate from these component measurements.

\paragraph{Motion regularization protocol}
\Cref{tab:icme_motion_ablation} separates encoding from motion supervision.
The origin control regresses a 3D axis point; the moment control predicts two axis-orthogonal coefficients.
The primary comparison fixes moment encoding and changes only $\lambda_M$; each flow uses targets re-encoded by its VAE.
Hard-joint motion RMSE is split by joint type. RMSE reuses the fixed anchors and normalized progress of the motion loss, averaging squared distances over anchors and states, then parts and objects before taking the square root; joint-type subsets use ground-truth labels.
Training pools decoded attributes using annotated part groups; public-benchmark generation receives neither shape nor part-count assistance.

\paragraph{Ablation results}
Without motion supervision, moment encoding improves all six metrics over origin encoding.
With moment encoding fixed, motion supervision reduces RMSE-R/P from 0.0431/0.0207 to 0.0362/0.0168 (16.0\%/18.8\%).
Their combination yields the lowest errors across all six metrics.

\subsection{Earlier component measurements}
The attribute-only baseline obtains 96.17\% type accuracy, 92.35\% axis accuracy, 94.54\% movable F1 and latent MSE 0.460 on 144 held-out objects ($n=3{,}987$ movable voxels).
Without kinematic-chain attention, the corresponding classification scores are 96.92\%, 93.35\% and 95.27\%.
These image-and-shape-conditioned scores use a supplied part count $K$; they are not measurements of the new motion regularizer.

\subsection{Appearance reconstruction}
\begin{table}[t]
\centering
\caption{Known-joint appearance reconstruction and diagnostics.}
\label{tab:icme_appearance}
\renewcommand{\arraystretch}{1.15}\small
\setlength{\tabcolsep}{4pt}
\begin{tabular}{@{}llcc@{}}
\toprule
Study & Setting & PSNR (dB) $\uparrow$ & IoU $\uparrow$\\
\midrule
Alignment & Before correction & 15.8 & 0.781\\
& After correction & 20.5 & 0.885\\
\midrule
Color source & Decoded & 19.72 & 0.872\\
& GT back-projected & 22.51 & 0.879\\
\bottomrule
\end{tabular}
\end{table}
A coordinate-alignment diagnostic shows the importance of accurate feature projection: correcting a $90^\circ$ frame mismatch improves PSNR and IoU (\cref{tab:icme_appearance}).
Over 20 objects, mean voxel-silhouette IoU increases from 0.617 to 0.876, with improvement on 19 objects.

To isolate appearance error, we fix Gaussian positions, orientations, scales, opacities and count, and replace only their colors with ground-truth back-projections.
PSNR improves by 2.79 dB while IoU changes by 0.007, locating a residual error in the appearance pathway.
Color substitution is a diagnostic reference, rather than an inference method.
\Cref{fig:icme_states} shows reconstruction across four configurations with appearance attached to the same parts.

\subsection{Limitations}
The motion model treats fixed, revolute and prismatic parts, assumes coordinate-aligned axes and evaluates part motions independently.
Training with annotated part groups does not resolve clustering errors or cross-grid assignment errors at inference.
Run-to-run variability is not quantified in these evaluations.
Density and friction remain stored attributes rather than calibrated physical predictions.

\section{Conclusion}
We presented MoRE, Motion-Regularized Encoding for articulated 3D generation.
An axis-orthogonal joint-line encoding removes the arbitrary along-axis origin, while a differentiable motion objective compares the behavior of reconstructed joints.
Image-conditioned functional latents and part-bound Gaussian appearance support generation followed by explicit articulation.
Controlled ablations show that motion supervision reduces anchor-motion error with either encoding, and the combined design gives the lowest errors across all six metrics.
On PartNet-Mobility and ACD, MoRE obtains the best articulated-state CD and CLIP scores among the compared methods; joint prediction also yields lower axis, origin and range errors than ArtiLatent, with comparable type accuracy.

\bibliographystyle{IEEEbib}
\bibliography{main}
\end{document}